\documentclass[letterpaper, 10 pt, conference]{ieeeconf}  

\IEEEoverridecommandlockouts                              

\usepackage{amsmath} 
\usepackage{amssymb}  
\usepackage{graphicx}
\usepackage{epstopdf}
\usepackage{amsmath}
\usepackage{mathtools}
\usepackage{dsfont}
\usepackage{tikz}
\usepackage{siunitx}
\usepackage{hyperref}

\usepackage[font=footnotesize]{caption}
\usepackage{subcaption}
\usepackage{floatrow}
\usepackage[noabbrev,capitalize]{cleveref}

\usepackage{balance}    
\usepackage{amsthm}
\usepackage{xcolor}
\usepackage{tcolorbox}
\usepackage{algorithm}
\usepackage{algpseudocode}
\usepackage{dsfont}

\usepackage[noadjust]{cite} 
\usepackage{amsmath,stmaryrd,graphicx}
\usepackage{mathtools}
\usepackage{placeins}

\newtheoremstyle{boldStyle}
  {\topsep}
  {\topsep}
  {\itshape}
  {0pt}
  {\bfseries}
  {.}
  { }
  {\thmname{#1}\thmnumber{ #2}\thmnote{ (#3)}}

\newtheoremstyle{italicStyle}
  {\topsep}
  {\topsep}
  {}
  {0pt}
  {\bfseries}
  {.}
  { }
  {\thmname{#1}\thmnumber{ #2}\thmnote{ (#3)}}

\theoremstyle{boldStyle}

\theoremstyle{italicStyle}

\newcommand{\BR}{\mathbb R}

\newcommand{\A}{\mathtt A}
\newcommand{\B}{\mathtt B}
\newcommand{\C}{\mathtt C}

\def\sq{\mathbin{{\strut\rule{1.25ex}{1.25ex}}}}

\definecolor{ugoColor}{rgb}{0.6,0.8,0.0}
\definecolor{ligthGray}{rgb}{0.95,0.95,0.95}

\makeatletter
\newcommand{\fixed@sra}{$\vrule height 2\fontdimen22\textfont2 width 0pt\shortrightarrow$}
\newcommand{\shortarrow}[1]{%
  \mathrel{\text{\rotatebox[origin=c]{\numexpr#1*45}{\fixed@sra}}}
}
\makeatother

\IEEEoverridecommandlockouts

\title{\LARGE \bf
Preparation of Papers for IEEE Sponsored Conferences \& Symposia*
}

\author{Albert Author$^{1}$ and Bernard D. Researcher$^{2}$
\thanks{*This work was not supported by any organization}
\thanks{$^{1}$Albert Author is with Faculty of Electrical Engineering, Mathematics and Computer Science,
        University of Twente, 7500 AE Enschede, The Netherlands
        {\tt\small albert.author@papercept.net}}%
\thanks{$^{2}$Bernard D. Researcheris with the Department of Electrical Engineering, Wright State University,
        Dayton, OH 45435, USA
        {\tt\small b.d.researcher@ieee.org}}%
}

\let\oldtwocolumn\twocolumn
\renewcommand\twocolumn[1][]{%
    \oldtwocolumn[{#1}{
    \begin{center}
    \centering
    \vspace{-15pt}
    \includegraphics[width=1\linewidth]{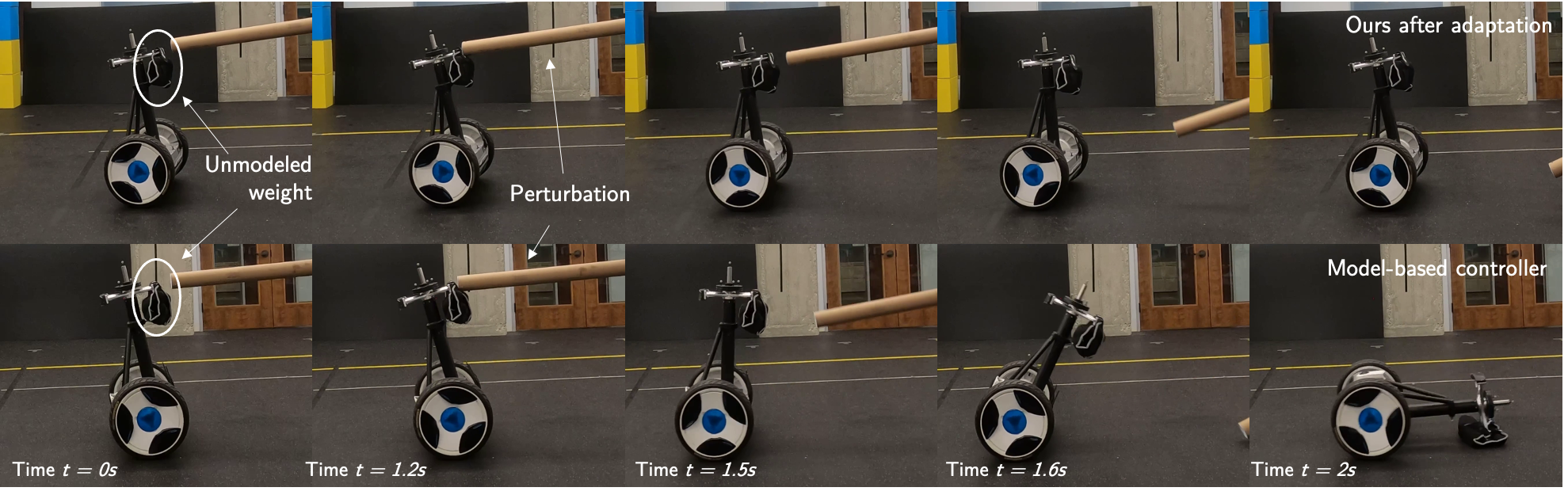}
    \captionof{figure}{Stabilizing a segway with an unmodelled weight against a force disturbance. \textbf{Top:} Our method is able to stabilize using a fully learned dynamics model that uses only 100 data points collected from one minute of data. \textbf{Bottom:} A state-of-the-art model predictive control policy using the nominal dynamics is unable to stabilize.}
    \label{fig:teaser}
    \end{center}
    }]
}

\begin{document}

\title{\LARGE \bf Learning to Control an Unstable System with One Minute of Data:\\ Leveraging Gaussian Process Differentiation in  Predictive Control  
\thanks{$^{*}$ This work was supported by NSF awards 1637598, 1645832, 1932091, 1924526, and 1923239, and funding from AeroVironment, JPL and BMW.}}

\author{
Ivan D. Jimenez Rodriguez, Ugo Rosolia, Aaron D. Ames, Yisong Yue
\thanks{
$^{1}$I. D. Jimenez Rodriguez, U. Rosolia, A. D. Ames and Yisong Yue are at the California Institute of Technology, Pasadena, USA.
E-mails: \tt\scriptsize{\{ivan.jimenez, urosolia, ames, yyue\}@caltech.edu.}}}



%

\maketitle
\thispagestyle{empty}
\pagestyle{empty}

\begin{abstract}
We present a straightforward and efficient way to control unstable robotic systems using an estimated dynamics model. Specifically, we show how to exploit the differentiability of Gaussian Processes to create a state-dependent linearized approximation of the true continuous dynamics that can be integrated with model predictive control. Our approach is compatible with most Gaussian process  approaches for system identification, and can learn an accurate model using modest amounts of training data.  
We validate our approach by learning the dynamics of an unstable system such as a segway with a 7-D state space and 2-D input space (using only one minute of data), and we show that the resulting controller is robust to unmodelled dynamics and disturbances, while state-of-the-art control methods based on nominal models can fail under small perturbations. Code is open sourced at \hyperref[https://github.com/learning-and-control/core]{https://github.com/learning-and-control/core}.
\end{abstract}

\IEEEpeerreviewmaketitle

\section{Introduction}


System identification is frequently used in robotics to mitigate model imperfections using measured input-output data \cite{ lai1982least, khosla_parameter_1985, chen1990non,caydas_support_2012,willis_systems_nodate}. Managing these modeling errors can be critical to achieving desired performance or guaranteeing safety. This problem is particularly challenging in systems with unstable dynamics since even small modeling errors can integrate over time without control inputs that can directly dampen them.  For instance, we experimentally show that running state-of-the-art model predictive control \cite{borrelli2017predictive} on a 7-D state space and 2-D input space segway system using a  misidentified model can lead to unsafe and unstable behavior, as depicted in Figure \ref{fig:teaser}.

A useful system identification framework must balance computation time, accuracy and data efficiency. Furthermore, since data often cannot be collected for the entire state space of a real system, an estimate of model uncertainty is also useful to plan around gaps in the knowledge of the learned model.  
Because of these challenges, much of contemporary research has focused on learning residuals of an already well-developed nominal dynamics model \cite{aswani2013provably, papadimitriou_control_2020, koller2018learning, rosolia2019learning, hewing2019cautious, kabzan2019learning, klenske2015gaussian,shi2019neural,shi2020neural,taylor2020learning, taylor2019episodic,chang2017learning, bujarbaruah2018adaptive}.


In this paper, we aim to learn the full dynamics models of unstable robotic systems.  Our goal is to develop a straightforward and data efficient method for system identification that can be easily integrated with state-of-the-art control methods.  We ground our approach in Gaussian processes (GPs), which are a popular method for learning dynamics models \cite{deisenroth_gaussian_2015,koller2018learning,hewing_cautious_2020,wang2018safe,kocijan2004gaussian,hewing2019cautious, kabzan2019learning, klenske2015gaussian,chang2017learning}.
We leverage the differentiability of GPs \cite{mchutchon2013differentiating, solak_derivative_nodate} to train a discrete-time dynamics model from training data of the form $(x_t, u_t, x_{t+\Delta t})$, while still recovering a state-dependent linearization of the dynamics that exploits the underlying continuous dynamics structure.

Learning a discrete-time linearizable dynamics has three key benefits. First, the approach can be very data efficient, as the differentiated GP model automatically infers the state-dependent linearization at every state. This differs from other approaches where the continuous dynamics model is learned directly and then used with collocation for approximation \cite{lee_gp-ilqg_2017}. As shown in \cite{levine2021framework}, learning the continuous dynamics rather than the discrete flow-map often requires higher sampling frequencies where measurement noise can become significant in practice. Second, one can use the estimated model with state-of-the-art model predictive control (MPC) methods \cite{borrelli2017predictive}  for effective and computationally efficient control synthesis that can handle state and input constraints. The final benefit is that the approach is generic and can be applied to many GP-based modeling approaches as a drop-in subroutine. 

The idea of using GPs for MPC is not new, but prior work either required using  computationally expensive procedures \cite{kocijan2004gaussian}, or limited themselves to learning only residual models \cite{hewing2019cautious, kabzan2019learning, klenske2015gaussian,chang2017learning}. Other prior work use of GPs with Dynamic Programming frameworks that do not take into account state and input constraints \cite{pan_probabilistic_nodate}. Also, unlike reinforcement learning approaches that use Policy Gradient \cite{deisenroth_pilco_nodate} or Value Iteration \cite{lutter_value_2021} with GPs, this method uses MPC to infer a policy. Learning the dynamics in MPC separates the policy from the dynamics and allows for using the same dynamics for different objectives.

\begin{figure*}[!h]
    \vspace{5pt}
    \includegraphics[width=1\linewidth]{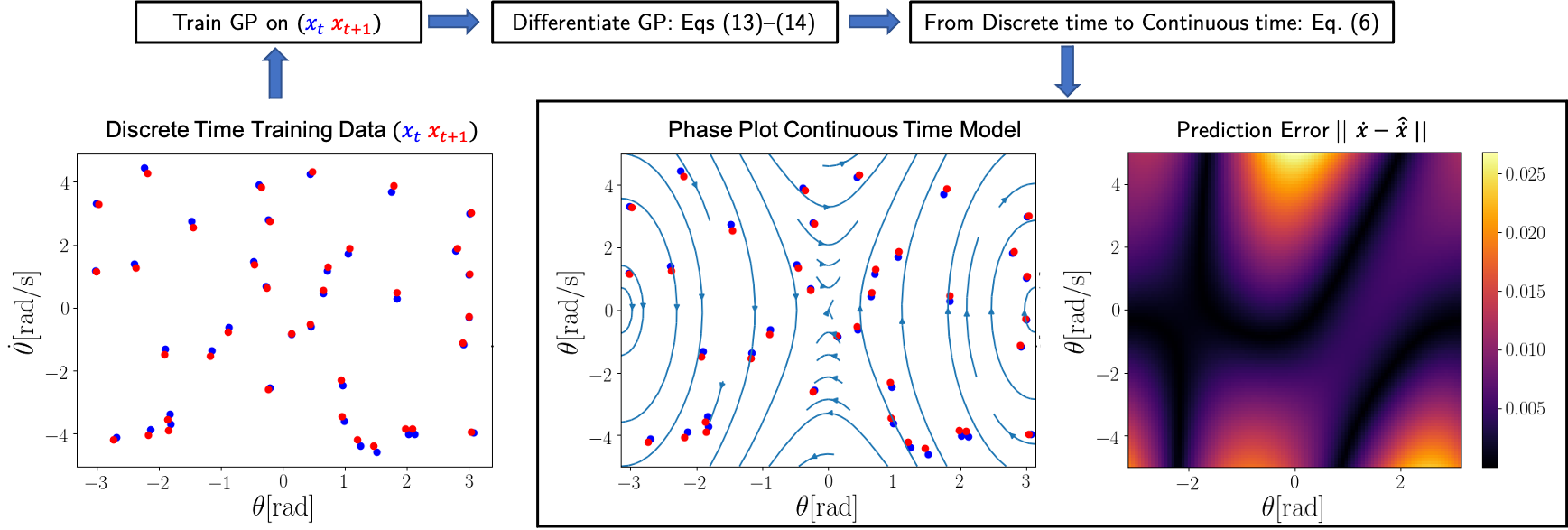}
    \caption{
    System identification results for a simulated pendulum.
    \textbf{Left-to-Right}: The dataset collected (selecting 30 initial points uniformly at random and integrating forward for $0.01s$);  Phase plot of estimated dynamics with dataset overlaid; Point-wise error between true and estimated dynamics.  
    Phase plots are computed on a $100 \times 100$ grid.
    We see that the error is small and captures the behavior of the system even in regions with few data points.
    \vspace{-15pt}
    }
    \label{fig:pend_ctn_learn}
\end{figure*}

We validate our approach by controlling unstable robotic systems, both in simulation and on a segway with 7-D state space and 2-D input space.
Whereas state-of-the-art control methods can fail to stabilize the segway under small model mismatch, we show that one can robustly stabilize using our model trained on only one minute of data (see Figure \ref{fig:teaser} above).
These results showcase the practical potential of our approach to significantly reduce the effort required for accurate system identification in unstable robotic systems.

\section{A Differentiation-based  Gaussian Process Model for Dynamics Estimation}

In this section, we outline our approach for system identification using a differentiation-based Gaussian process  model.  We first describe and motivate learning a state-dependent linearized model in Section \ref{sec:sysid}.  We then discuss Gaussian process preliminaries in Section \ref{sec:gp}, and how to differentiate a GP to obtain a state-dependent linearized dynamics model in Section \ref{sec:diffGP}.

\subsection{State-Dependent Linearized Dynamics Modeling} 
\label{sec:sysid}
 
We consider the following dynamical system:
\begin{align}
    \dot{x}(t) = f(x(t),u(t)),
    \label{eq:dyn}
\end{align}
 with states $x \in \BR^n$, control inputs $ u \in \BR^m$, and where $f$ is unknown (not even the functional form). The above system is subject to the the following state and input constraints:
 \begin{equation}\label{eq:constraints}
     x(t) \in \mathcal{X} \text{ and }u(t)\in \mathcal{U},~\forall t \geq 0.
 \end{equation}
 Our goal is to design a control policy $\pi:\mathbb{R}^n \rightarrow \mathbb{R}^d$ which maps states to actions. In order to compute such a policy we will use historical data to estimate the full system dynamics $f$, which will be leveraged in a predictive control scheme.  We are particularly interested in systems that are passively unstable (e.g., a segway).
 
 We assume access to a dataset of $M$ state-input pairs $\{\hat{x}(iT), u(iT)\}_{i = 0}^M$, where $T$ is the sampling period. Visually, on the pendulum, this would correspond to the left-most plot in \cref{fig:pend_ctn_learn}. Furthermore, we assume the control action is applied using a sampling-and-hold strategy meaning that:
 \begin{align}
     \hat{x}((i+1)T) = \int_{iT}^{(i+1)T} f(x(\tau), u(t)) d \tau + x(t)  + w,
     \label{eq:flow_map}
 \end{align}
 where the noise $w$ a zero-mean Gaussian, i.e, $w \sim \mathcal{N}(0, \sigma^2)$.
 

 Rather than performing motion planning directly on these discrete-time dynamics (as is common in the literature \cite{deisenroth_gaussian_2015,hewing_learning-based_2020}), we instead use  \eqref{eq:flow_map} and a state-dependent linear approximation of the dynamics around the state-input pair $(\bar x, \bar u)$:
 \begin{equation}
 \begin{aligned}
     f(x,u) &\approx A(\bar{x}, \bar{u})x + B(\bar{x}, \bar{u})u + C(\bar{x}, \bar{u}), \\
     A(\bar{x}, \bar{u}) &= \frac{\partial f(\bar{x}, \bar{u})}{\partial x}, \\
     B(\bar{x}, \bar{u}) &= \frac{\partial f(\bar{x}, \bar{u})}{\partial u }, \\
     C(\bar{x}, \bar{u}) &= f(\bar{x}, \bar{u}) - \frac{\partial f(\bar{x}, \bar{u})}{\partial x} \bar{x} - \frac{\partial f(\bar{x}, \bar{u})}{\partial u } \bar{u}.
 \end{aligned}
 \label{eq:ctn_lin_approx}
 \end{equation}
 This linearization of the dynamics can be related to a linearization of the discrete flow map in \eqref{eq:flow_map} as follows:
 \begin{equation}
     \begin{aligned}
      \hat{x}(t + \delta t)                     &\approx \A(\bar{x}, \bar{u}) x(t) + \B(\bar{x}, \bar{u}) u_t + \C(\bar{x}, \bar{u}), \\ 
    \A(\bar{x}, \bar{u}) &= e^{A(\bar{x}, \bar{u}) \delta t}, \\
    \B(\bar{x}, \bar{u}) &= A(\bar{x}, \bar{u})^{-1}(e^{A(\bar{x}, \bar{u}) \delta t} - I)B(\bar{x}, \bar{u}), \\
    \C(\bar{x}, \bar{u}) &= A(\bar{x}, \bar{u})^{-1}(e^{A(\bar{x}, \bar{u}) \delta t} - I)C(\bar{x}, \bar{u}).
     \end{aligned}
     \label{eq:disc_lin_approx}
 \end{equation}
 Although these matrices alone are sufficient for use in our MPC controller, through the use of matrix logarithms, the local linear approximation of the dynamics can be computed:
\begin{equation}
    \begin{aligned}
        A(\bar{x},\bar{u}) &\approx \frac{1}{\delta t} \log(\A(\bar{x},\bar{u})), \\
        B(\bar{x},\bar{u}) &\approx  \A(\bar{x},\bar{u})^{-1} A(\bar{x},\bar{u}) \B(\bar{x},\bar{u}),\\
        C(\bar{x},\bar{u}) &\approx \A(\bar{x},\bar{u})^{-1} A(\bar{x},\bar{u}) \C(\bar{x},\bar{u}).
    \end{aligned}
    \label{eq:disc_to_ctn_lin_approx}
\end{equation}

 As we shall see in \cref{sec:control}, having a state-dependent linearization is crucial for efficient integration with predictive control.
 In general, computing a state-dependent linearization with GPs in real-time can be challenging which is why most prior work resorts to approximating the GP using inducing inputs \cite{hewing_learning-based_2020, quinonero2005unifying}, a time-varying state-input independent model \cite{klenske2015gaussian} or learning the residual \cite{hewing2019cautious, kabzan2019learning, klenske2015gaussian}. We will show in Section \cref{sec:diffGP} how to solve for the matrices of the state-dependent linear dynamical approximation, $(A,B,C)$, by taking derivatives of a Gaussian process dynamics model.


\subsection{Gaussian Process Preliminaries}
\label{sec:gp}
A Gaussian process (GP) is the defined by a mean function $\mu(s)$ and positive semidefinite covariance function $k(s, s')$. In this work we will primarily use two kernels: the Radial Basis Function (RBF) Kernel:
\begin{align}
    k_{rbf}(s,s') = \exp\left(-\frac{\lVert s - s' \rVert_2^2}{2 \sigma^2}\right),
\end{align}
and the Periodic Kernel:
\begin{align}
    k_{p}(s,s') = \exp\left(-\frac{2 \sin^2(\frac{\pi \lvert s - s' \rvert }{\omega 2})}{\ell^2})\right),
\end{align}
 where $\sigma$, $\omega$ and $\ell$ are tunable parameters. We will also construct composite kernels by exploiting the fact that the product of kernels is a valid kernel, in order to encode geometric properties of the dynamical system.  In particular, the Periodic Kernel is useful for modeling angular coordinates, whereas the RBF kernel is more suitable for Euclidean coordinates.
 
Samples of a GP take the form:
\begin{align}
 h \sim \mathcal{GP}\left(\mu\left(\begin{bmatrix}x \\ u \end{bmatrix}\right), k\left(\begin{bmatrix}x \\ u \end{bmatrix}, \begin{bmatrix}x' \\ u' \end{bmatrix} \right)\right),
 \label{eq:dyn_gp}
\end{align}
where function samples approximate the integral of the dynamics as follows:
\begin{align}
 h\left(\begin{bmatrix}x(t) \\ u(t) \end{bmatrix}\right) \approx \int_t^{t+ \delta t} f(x(\tau), u(\tau))    d \tau.
\end{align}
Multi-dimensional outputs are predicted with an independent GP for each output.
For a test input $s = \begin{bmatrix}\tilde{x} & \tilde{u}\end{bmatrix}^\top$, the mean and variance are computed as:
\begin{align}
    \mu\left(  s \right) &= \mathbb{E}_h\left[ h\left( s \right) \right] = k\left(s, \mathcal{X} \right)(k(\mathcal{X}, \mathcal{X}) + \sigma_\epsilon I )^{-1} \mathcal{Y},  \nonumber\\
    \sigma_h^2 &= \mathbb{V}_f \left[ h\left( s\right)\right] \nonumber\\
    & =  k\left(s, s\right) - k \left(s, \mathcal{X}\right) (k(\mathcal{X}, \mathcal{X}) + \sigma_\epsilon I )^{-1} k \left(\mathcal{X}, s\right).\nonumber
\end{align}

\subsection{Differentiating a Gaussian Process}\label{sec:diffGP}

From \cite{mchutchon2013differentiating, solak_derivative_nodate}, we know that the derivative of a GP is another GP. For the following derivation it is sufficient for the kernel function to be differentiable with respect to the both of its parameters (which is true for both the RBF and Periodic kernels). For an input $s$ we define the derivative of a GP as follows:
\begin{align}
    h' \sim \mathcal{GP}(\mu', k'(s,s')),
\end{align}
where $h': \BR^{n+m} \rightarrow \BR^{n+m}$ is the gradient of sampled function $h$. We derive the mean of the GP derivative as follows:
\begin{equation}
\begin{aligned}
    \mu'(S) &= \mathbb{E}\left[ \frac{\partial}{\partial s} h \right]= \frac{\partial}{\partial s} \mathbb{E}\left[h \right] = \frac{\partial}{\partial s} \mu(s),
\end{aligned}
\end{equation}

This GP in \eqref{eq:dyn_gp} is related to the linear approximation in \eqref{eq:disc_lin_approx} as follows:
\begin{equation}\label{eq:ABC_GP}
    \begin{aligned}
        \A(\bar{x},\bar{u}) &\approx \frac{\partial h(\bar{x}, \bar{u})}{\partial x}  + I, \\
        \B(\bar{x},\bar{u}) &\approx \frac{\partial h(\bar{x}, \bar{u})}{\partial u}, \\
        \C(\bar{x},\bar{u}) &\approx h(\bar{x}, \bar{u}) + \bar{x} - \A(\bar{x},\bar{u})\bar{x} - \B(\bar{x},\bar{u})\bar{u},
    \end{aligned}
\end{equation}
These approximates are derived by noting that the GP approximates the integral in \eqref{eq:flow_map} which concludes the derivation. 

Graphically, this relationship can be seen in the center plot of \cref{fig:pend_ctn_learn} where the data points are overlapped with the state-dependent continuous linear approximation we computed using \cref{eq:disc_to_ctn_lin_approx}.
\begin{algorithm}[b]
  \caption{Control Policy}\label{euclid}\label{algo:policy}
  \begin{algorithmic}[1]
      \State \textbf{Init Parameters:} ${\boldsymbol{x}}^*_{t-1}$, ${\boldsymbol{u}}^*_{t-1}$, $\mathcal{GP}(\mu'(s) , k'(s, s'))$
      \State \textbf{Input:} $x_t$
      \If {$t>0$}\Comment{Update Candidate Trajectory}
      \State Set $\bar x_{k|t} = x_{k|t-1}^*,~\forall k\{t,\ldots,t+N-1\}$
      \State Set $\bar u_{k|t} = u_{k|t-1}^*,~\forall k\{t,\ldots,t+N-2\}$
      \State Set $\bar x_{t+N|t} = x_{t+N-1|t-1}^*$
      \State Set $\bar u_{t+N-1|t} = u_{t+N-2|t-1}^*$
      \EndIf
      \For{$k \in \{t, \ldots, t+N-1\}$} \Comment{Update Model}
        \State Set $A_k = \A(\bar x_{k|t},\bar u_{k|t})$ from expectation of ~\eqref{eq:ABC_GP}
        \State Set $B_k = \B(\bar x_{k|t},\bar u_{k|t})$ from expectation of ~\eqref{eq:ABC_GP}
        \State Set $C_k = \C(\bar x_{k|t},\bar u_{k|t})$ from expectation of ~\eqref{eq:ABC_GP}
      \EndFor
      \State Solve the FTOCP~\eqref{eq:ftocp} with $\{A_k,B_k,C_k\}_{k=t}^{t+N-1}$
      \State Store $\boldsymbol{x}^*_t = [x_{t|t}^*, \ldots, x_{t+N|t}^*]$
      \State Store $\boldsymbol{u}^*_t = [u_{t|t}^*, \ldots, u_{t+N-1|t}^*]$
      \State Set $u_t = u_{t|t}^*$
      \State \textbf{Outputs} $u_t$
  \end{algorithmic}
\end{algorithm}
Note that since we are training on M, a dataset of state input pairs, we are still learning the discrete time flow map shown in \eqref{eq:flow_map}. A key aspect of our contribution is to re-interpret the learned dynamics in the context of \eqref{eq:dyn} to directly infer a local linear approximation that is amenable to MPC. This allows our method to be retroactively applied to previous GP-based modeling work that learns a discrete transition model.

\section{Control Design}\label{sec:control}

We now describe our control strategy, which is based on state-of-the-art methods for model predictive control (MPC) \cite{borrelli2017predictive}.  This approach works generically with the state-dependent linearized model described in \eqref{eq:disc_lin_approx}.
First, we introduce a Finite Time Optimal Control Problem (FTOCP) which is based on a simplified Affine Time-Varying (ATV) model. 
We then present the proposed algorithm that at each time $t$ solves an FTOCP where the ATV model is updated leveraging the state-dependent linearized model from \eqref{eq:disc_lin_approx}. Our algorithm applies the first action of the planned trajectory to the system and the entire process is repeated at the next time step $t+1$, yielding to a receding horizon strategy also referred to as model predictive control. 
\subsection{Finite Time Optimal Control}
At time $t$ and for system's state $x(t)$, we define the following finite time optimal control problem (FTOCP):
\begin{equation}\label{eq:ftocp}
    \begin{aligned}
        J(x(t)) = \quad & \min_{\boldsymbol{u}_t}  \sum_{k=t}^{t+N-1} l(x_{k|t}, u_{k|t}) + V(x_{t+N|t}), \\
         \text{s.t. } \quad & x_{k+1|t} = A_k x_{k|t} + B_k u_{k|t} + C_k, \\
        & x_{k|t} \in \mathcal{X}, u_{k|t} \in \mathcal{U}, \forall k \in \{t, \ldots, t+N\}\\
         & x_{t|t} = x(t)
    \end{aligned}
\end{equation}
where $\boldsymbol{u}_t = [u_{t|t}, \ldots, u_{t+N|t}]$ is a sequence of open-loop control actions, the stage cost $l:\mathbb{R}^n\times\mathbb{R}^d \rightarrow \mathbb{R}$, the terminal cost $V: \mathbb{R}^n \rightarrow \mathbb{R}$ and the sets $\mathcal{X} \subseteq \mathbb{R}^n$ and $\mathcal{U}\subseteq \mathbb{R}^d$ represent the state and input constraints, respectively. In the above problem, the time varying matrices $A_k$, $B_k$ and $C_k$ define a discrete time ATV approximation of the true system~\eqref{eq:dyn} and we will compute them using the differentiable GP from Section~\ref{sec:diffGP}.

We denote the optimal state-input sequences to the FTOCP~\eqref{eq:ftocp} as:
\begin{equation}\label{eq:optTrajectory}
\begin{aligned}
    \left( \boldsymbol{x}_t^*, \boldsymbol{u}_t^* \right)=\left[\left( x_{t|t}^*, u_{t|t}^* \right), \ldots, \left( x_{t+N|t}^*, u_{t+N-1|t}^* \right)\right]
\end{aligned}
\end{equation}
which minimize the predicted cost while satisfying state and input constraints from~\eqref{eq:constraints}. When the above optimal predicted trajectory is computed at time $t$, we have that $x_{k|t}^*$ denotes the predicted state of at time $k$. This notation will be useful later on when we are going to differentiate between the optimal state $x_{k|t}^*$ at time $k$ predicted at time $t$ and the optimal state $x_{k|t+1}^*$ at time $k$ predicted at time $t+1$. In what follows, we use the optimal state-input sequences~\eqref{eq:optTrajectory} to synthesize a control policy for the dynamical system~\eqref{eq:dyn}.

\subsection{Policy Synthesis}
This section describes the control synthesis strategy. At each time $t$, we solve the FTOCP~\eqref{eq:ftocp}, where the time varying matrices $\{A_k, B_k, C_k\}_{k=t}^{t+N-1}$ are computed using the differentiable GP evaluated along the candidate state input sequences:
\begin{equation*}
\begin{aligned}
    \left(\bar{\boldsymbol{x}}_t, \bar{\boldsymbol{u}}_t \right) = \left[ \left(\bar x_{t|t}, \bar u_{t|t} \right), \ldots \left( \bar x_{t+N|t}, \bar u_{t+N-1|t}\right) \right]
\end{aligned}
\end{equation*}
At time $t=0$, we initialize the candidate trajectory with an initial guess and afterwards we update the candidate solution using the optimal trajectory from~\eqref{eq:optTrajectory}, as shown in \cref{algo:policy}. In particular, in \cref{algo:policy} we update the candidate trajectory by shifting the optimal solution computed as the previous time time~(Lines~3-8). Afterwards, we update ATV matrices used to define the FTOCP problem~\eqref{eq:ftocp}. Finally, we solve problem~\eqref{eq:ftocp} and we store the optimal state-input trajectories. The strategy described in Algorithm~\ref{algo:policy} is repeated at each time $t$ based on the new measurement $x_t$.

It is clear that the prediction model defined by the ATV matrices $\{A_k, B_k, C_k\}_{k=t}^{t+N-1}$ plays a crucial role in determining the success of the MPC. If the prediction model is inaccurate, then the closed-loop system will deviate from the planned trajectory. This deviation may result in poor closed-loop performance and safety constraint violation. 
We validate this point in our experiments showing that controlling using an inaccurate model can be unsafe, thus highlighting the need to quickly learn accurate dynamics models.


\begin{figure}[b!]
    \begin{subfigure}{0.47\linewidth}
    \includegraphics[trim =3mm 3mm 3mm 3mm, width=1.0\linewidth]{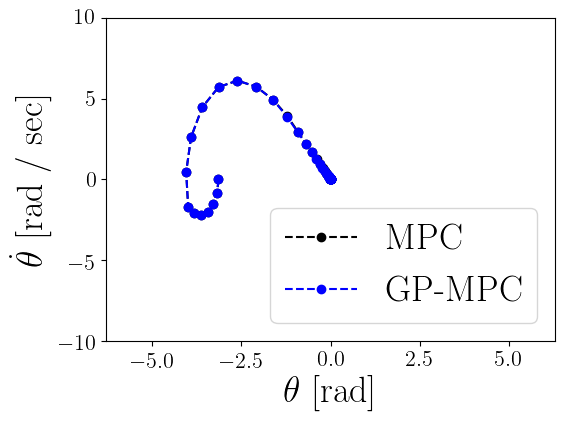}
    \vspace{-18pt}
    \caption{}
    \label{fig:pendulum_traj}
    \end{subfigure}
    \quad%
    \begin{subfigure}{0.47\linewidth}
    \includegraphics[trim =3mm 3mm 3mm 3mm, width=1.0\linewidth]{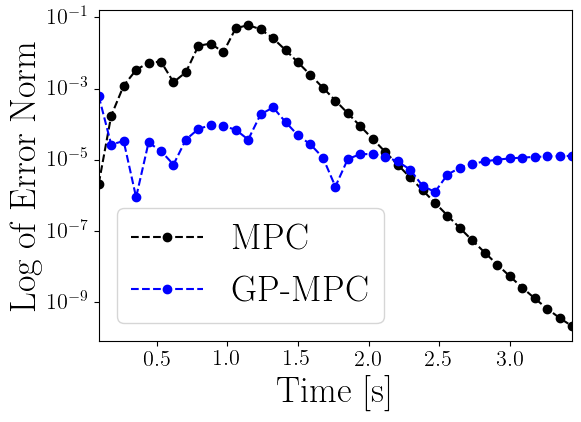}
    \vspace{-18pt}
    \caption{}
    \label{fig:pendulum_one_step}
    \end{subfigure}
    \label{fig:pendulum_sim}
    \vspace{-7pt}
    \caption{\cref{fig:pendulum_traj} shows the closed-loop trajectory of the pendulum. MPC with both the true model and the GP produce near identical trajectories. \cref{fig:pendulum_one_step} shows the one-step prediction error of the MPC policy for both models.}
\end{figure}

\begin{figure*}[t]
    \vspace{5pt}
    \begin{subfigure}{0.3\linewidth}
    \includegraphics[width=1.0\linewidth]{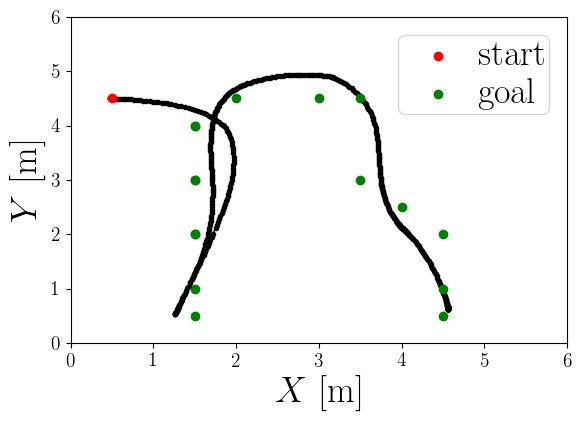}
    \vspace{-20pt}
    \caption{}
    \label{fig:sim_segway_to_goal}
    \end{subfigure}
    \quad%
    \begin{subfigure}{0.3\linewidth}
    \includegraphics[width=1.0\linewidth]{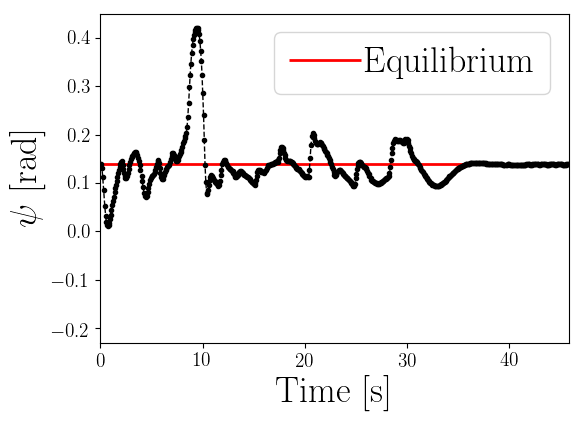}
    \vspace{-20pt}
    \caption{}
    \label{fig:sim_segway_balanced}
    \end{subfigure}
    \quad%
    \begin{subfigure}{0.3\linewidth}
    \includegraphics[width=1.0\linewidth]{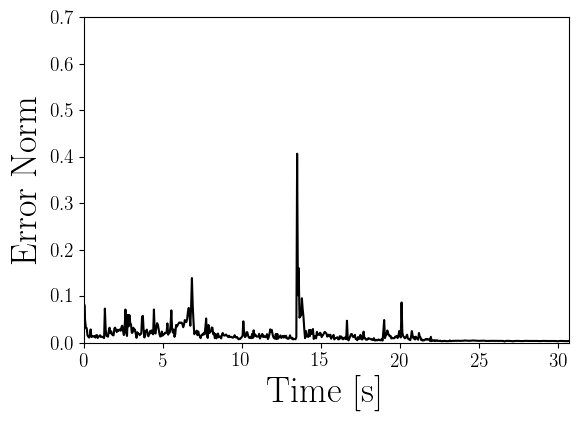}
    \vspace{-20pt}
    \caption{}
    \label{fig:sim_segway_to_goal_error_norm}
    \end{subfigure}
    \label{fig:sim_segway_plots}
    \vspace{-7pt}
    \caption{Control results on simulated segway.  \cref{fig:sim_segway_to_goal}, we plot the position of the segway as it reaches a sequence of target goals. \cref{fig:sim_segway_balanced} shows $\psi$: the angle of the segway with respect to the upright position. Notice that the segway must deviate from its equilibrium in order to accelerate forwards or backwards. \cref{fig:sim_segway_to_goal_error_norm} shows the one-step prediction error of the MPC policy using GP dynamics. 
    \vspace{-20pt}}
\end{figure*}%

\section{Simulation Results}

In this section, we validate our approach in simulation on two unstable systems: the inverted pendulum and the segway.  The goal of this evaluation is to provide both an intuition as well as a demonstration of the theoretical  limits for our approach.  
Specifically, we aim to address:
\begin{itemize}
    \item Can we learn an accurate dynamics model with few training examples?
    \item Can we integrate our dynamics model with control for steering and motion planning?
\end{itemize}


\subsection{Pendulum Simulation}
We first test on a continuous time inverted pendulum. The state of the system is $x = [ \theta, \dot{\theta}]$ with torque input $u$. The system has a point mass of $0.25$kg and a length of $0.5$m. 

\subsubsection{Data Collection \& System Identification}
First, we estimate the discrete flow map for the unactuated system using 37 samples collected uniformly at random so that $\theta(t_i) \sim U(-\pi, \pi)$ and $\dot{\theta}(t_i) \sim U(-5, 5)$. We capture the geometric structure of the pendulum's sate-space by using the following kernel:
\begin{equation}
    \begin{aligned}
    k(x, x') = k_p(\theta, \theta')k_{rbf}(\dot{\theta}, \dot{\theta}'),
    \end{aligned}
    \label{eq:pendulum_kernel}
\end{equation}
since $x \in \mathbb{S}^1 \times \mathbb{R}$. Next, we use a local linear approximation to compute a point-wise estimate of the continuous dynamics as shown in~\eqref{eq:ABC_GP}. The entire dataset of state transitions is shown in the left plot of \cref{fig:pend_ctn_learn}.

\subsubsection{Evaluating Model Accuracy}
We divide the state-space of the pendulum into a $100 \times 100$ grid. For each point in the grid we compute the true value of $\dot{x}$ using~\eqref{eq:dyn} and an estimated value using the linear approximation in~\eqref{eq:ctn_lin_approx} where $A$, $B$ and $C$ are computed using the derivative of the GP as in~\eqref{eq:ABC_GP}. The phase plot in \cref{fig:pend_ctn_learn} correspond to the direction of the estimated $\dot{x}$. For each point in the grid we can compute the 2-norm of the error which is shown in the heat map on the right.  Overall, we can see that our approach can recover an accurate state-dependent linearization of the true continuous time dynamics.

\subsubsection{Evaluating Motion Planning}
Next, we consider the task of steering the pendulum to the upright position starting from $x(0) = (-\pi, 0)$ with the input constraint set $\mathcal{U}=\{ u ~ | ~ -.6 \leq u \leq .6 \}$. Notice that with these constraints the pendulum is unable to reach the goal in a single swing. For this task, we collect a new dataset of 34 uniformly distributed samples where states are sampled as before and $u(t_i) \sim U(-.6, .6)$. Finally, we run the control policy from \cref{algo:policy}. Both our strategy and an MPC using true dynamics are able to swing up the pendulum reaching the unstable equilibrium state as shown in \cref{fig:pendulum_traj}.


To evaluate the performance of the GP model for motion planning, we compute the difference between the first state predicted by the MPC policy and the actual state observed by the system.  \cref{fig:pendulum_one_step} shows the errors for the true model and the learned model at each time step of the pendulum's swing-up. We see that throughout the pendulum swing up the MPC model has a higher prediction error than the GP. Towards the end, as the pendulum stabilizes, the error of the MPC policy with the true model falls to 0 while the GP controller maintains a low but stable error. We expect the MPC with continuous time dynamics to have a slightly higher error than the GP as
the continuous time dynamics are linearized and discretized to compute the predicted trajectory, while GP provides an estimate of the discrete-time map which is used to compute the next step.

\subsection{High-Fidelity Segway Simulation}
\label{sec:segway_sim}
We next evaluate our strategy on a high-fidelity simulated segway based on the 6-D state space and 2-D input space system shown in Figure \ref{fig:highFidSim}.
The state of a segway is $x= [X, Y, \theta, v, \dot \theta, \psi, \dot \psi]$, where $(X,Y)$ represents the position of the center of mass, $(\theta, \dot \theta)$ the heading angle and yaw rate, $v$ the velocity and $(\psi, \dot \psi)$ the rod's angle and angular velocity. The control input $u = [T_l, T_r]$, where $T_l$ and $T_r$ are the torques to the left and right wheel motors, respectively. For all experiments, we limit $|T_l| \leq 6$ and $| T_r | \leq 6$.

\subsubsection{Data Collection \& System Identification}
Recall that we are learning the mapping shown in \eqref{eq:disc_lin_approx}. As a prior, we know that $X$ and $Y$ have no effect on the dynamics of the system. Therefore we only need to learn a mapping from the state excluding $X$ and $Y$ at the current time step to the change in state at the next time step. We encode the property of $\theta$ being an angle via the following kernel:
\begin{equation}    
    \begin{aligned}
    k(x,x') &= k_p(\theta, \theta')k_{rbf}(s,s'),
    \end{aligned}
\label{eq:segway_learned_function}
\end{equation}
where $s = [v, \dot \theta, \psi, \dot \psi]$. Although $\psi$ is also an angle, since the system cannot rotate about that axis without catastrophic failure, we use a regular $RBF$ kernel for it.


In simulation, we record the segway performing a task consisting of $1000$ state-transitions at a frequency $T=0.05$ which is approximately one minute of data. We then find $180$ clusters using a hierarchical clustering algorithm and select the nearest neighbor for each cluster as the data-point for the training set. We test the ability of the our strategy to perform the same task that was used to collect the data but with the GP dynamics model.

\subsubsection{Evaluating Motion Planning}
\cref{fig:sim_segway_to_goal} shows the path that the segway takes while reaching the targets. Once the segway is within $1$m of a goal the next one is provided. Notice that peaks and troughs in \cref{fig:sim_segway_balanced} correspond to moments of forwards and backwards acceleration (since the segway must tilt to move forward). Those same moments of high acceleration also match with the peaks of high one-step prediction error observed in \cref{fig:sim_segway_to_goal_error_norm}.



\section{Experimental Results on Segway}
\label{sec:hardware}

\begin{figure}[t]
    \vspace{5pt}
    \centering
	\includegraphics[width= 0.4\columnwidth]{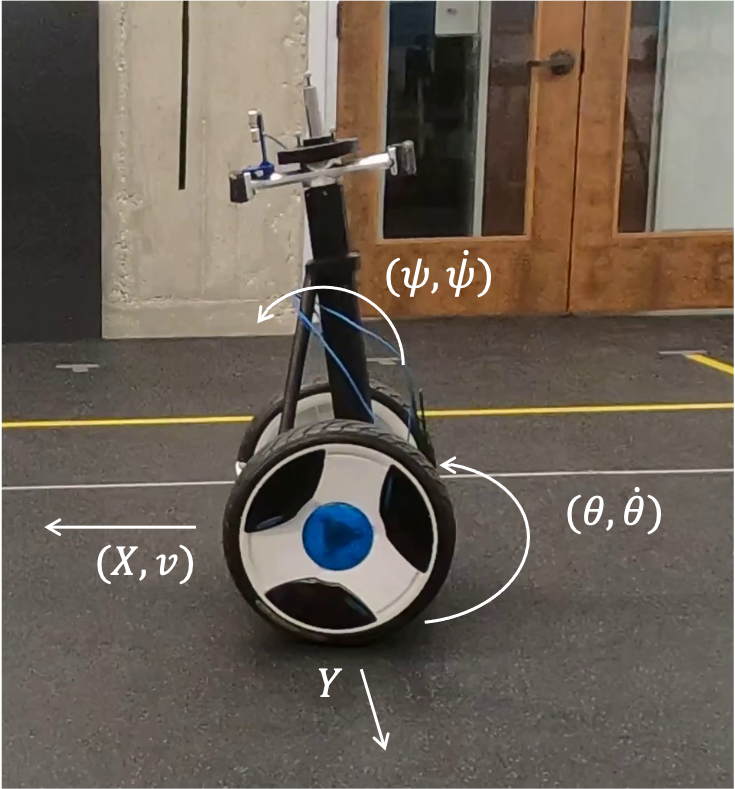}
    \caption{Experimental platform. \vspace{-20pt}}
    \label{fig:highFidSim}
\end{figure}

We finally evaluate our approach experimentally on the segway system (see Figure \ref{fig:highFidSim}). The state representation is the same as in \cref{sec:segway_sim}.

We aim to demonstrate that:

\begin{itemize}
    \item Our method can control a physical open-loop unstable system to perform a simple move-forward task.
    \item Our method is able to overcome perturbations with  unmodelled dynamics in a physical open-loop unstable system where a state-of-the art MPC controller fails.
\end{itemize}

\subsubsection{Data Collection \& System Identification}
We record the trajectory of the segway performing a task that takes $1000$ measurements recorded every $T=0.05s$ to complete. The data is then preprocessed using the same procedure as for the simulated system.
We evaluate on two tasks: the first is a standard moving forward task while staying upright, and the second is stabilizing task under a external force disturbance. We note that since this a real system an estimation is performed online with on-board sensors there is significant estimation error as well. Although our method is capable of running at  $20$Hz with up to 300 data-points, we required less data than this for the experiments.

\subsubsection{Simple Move-Forward Task} 
\begin{figure*}[h!]
    \vspace{5pt}
    \begin{subfigure}{0.32\linewidth}
    \hspace*{-10pt}
    \includegraphics[width=1.0\linewidth]{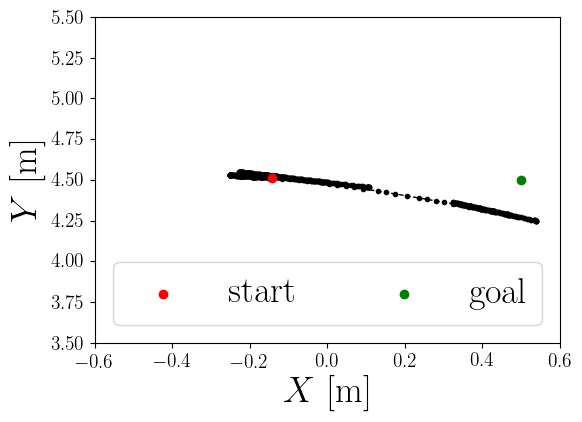}
    \vspace{-5pt}
    \caption{}
    \label{fig:hw_segway_to_point_position}
    \end{subfigure}
    \hspace*{-10pt}
    \begin{subfigure}{0.32\linewidth}
    \hspace*{-8pt}
    \includegraphics[width=1.0\linewidth]{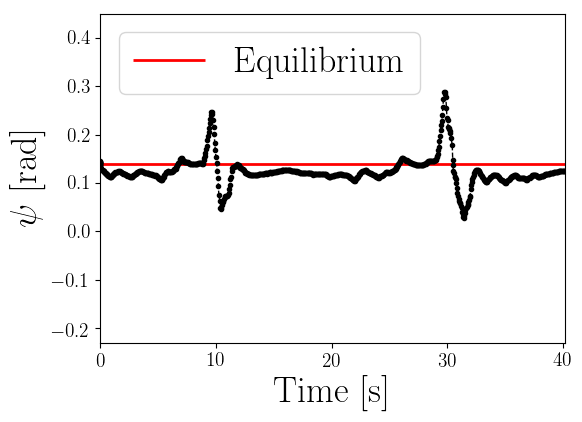}
    \vspace{-5pt}
    \caption{}
    \label{fig:hw_segway_balanced}
    \end{subfigure}
    \hspace*{-10pt}
    \begin{subfigure}{0.32\linewidth}
    \hspace*{-8pt}
    \includegraphics[width=1.0\linewidth]{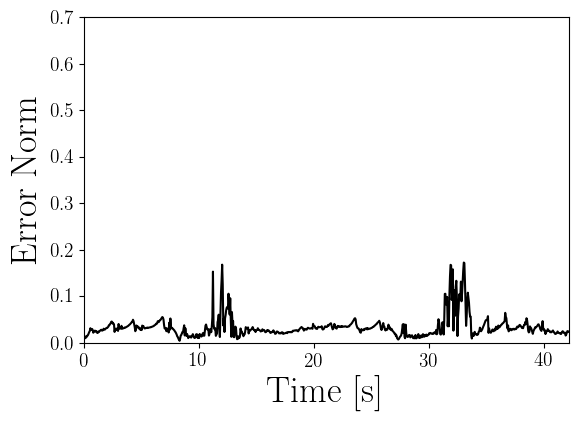}
    \vspace{-5pt}
    \caption{}
    \label{fig:hw_segway_to_goal_error_norm}
    \end{subfigure}
    \vspace{-5pt}
    \caption{The 7-D state space and 2-D input space segway going between two points.  In \cref{fig:hw_segway_to_point_position} we plot the physical position of the segway. In \cref{fig:hw_segway_balanced} we see $\psi$: the angle between segway's pole and the upright position. The two peaks correspond to the segway accelerating and decelerating. \cref{fig:hw_segway_to_goal_error_norm} shows the MPC policy's one-step prediction error using the GP dynamics.}
    \label{fig:hw_segway_to_point}
\end{figure*}

\begin{figure*}[!h]
\includegraphics[width=\linewidth]{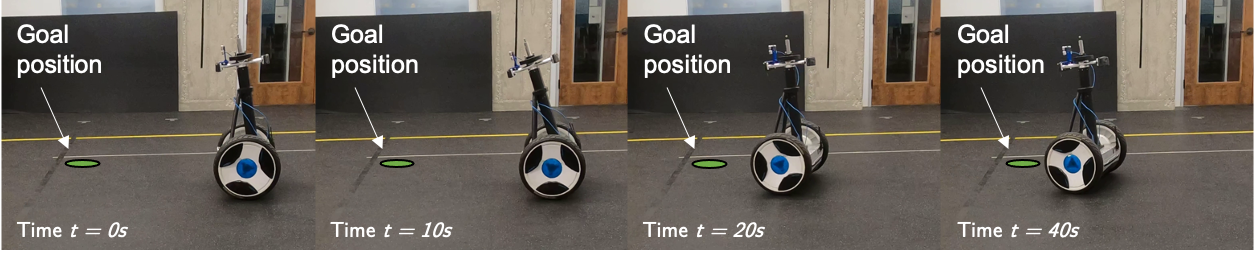}
    \caption{This image sequence shows the segway going to a point using the full order dynamics learned with our GP method.}
    \label{fig:hw_segway_to_point_seq}
\end{figure*}

We start by considering the move-forward task. For this we only require 130 data points. As can be observed in \cref{fig:hw_segway_to_point_seq}, the system is able to stabilize at a point, move forward and stabilize close to the new location with some minor oscillations around the target point, as highlighted in \cref{fig:hw_segway_to_point}. The first and second peaks in \cref{fig:hw_segway_balanced} correspond to the acceleration and deceleration respectively. Notice that due to a combination of modeling and estimation estimation error, the segway balances slightly off the equilibrium point. Finally, from \cref{fig:hw_segway_to_goal_error_norm} we see that the model error spikes at the moments of high acceleration.

\subsubsection{Robustness To Perturbations} We now evaluate the performance of the learned model under perturbations. To test the robustness of the learned model, we start by collecting 100 data points from a dataset of the MPC policy with nominal dynamics completing the task with an unmodelled weight of 2kg. Although this results in slightly different behavior, the amount of data collected is the same as in previous experiments. Next, we attach $4$kg of unmodelled weight as shown in \cref{fig:teaser}. Notice that the weight is not perfectly centered and that it is allowed to sway back and forth from its point of contact.

\begin{figure*}[h!]
    \begin{subfigure}{0.32\linewidth}
    \hspace*{-10pt}
    \includegraphics[width=1.0\linewidth]{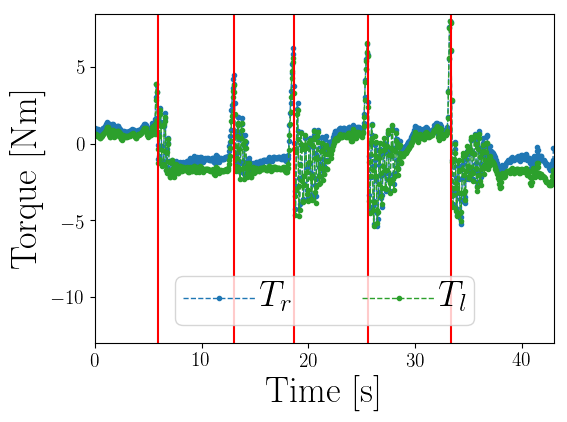}
    \vspace{-5pt}
    \caption{}
    \label{fig:hw_gp_perturbation_inputs}
    \end{subfigure}
    \hspace*{-10pt}
    \begin{subfigure}{0.32\linewidth}
    \hspace*{-10pt}
    \includegraphics[width=1.0\linewidth]{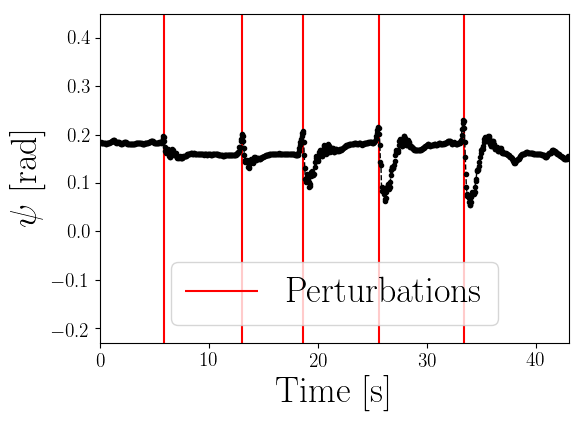}
    \vspace{-5pt}
    \caption{}
    \label{fig:hw_gp_perturbation_balanced}
    \end{subfigure}
    \hspace*{-10pt}
    \begin{subfigure}{0.32\linewidth}
    \hspace*{-10pt}
    \includegraphics[width=1.0\linewidth]{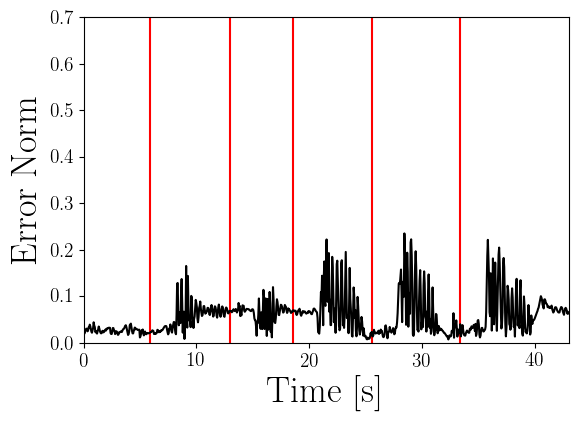}
    \vspace{-5pt}
    \caption{}
    \label{fig:hw_gp_perturbation_error_norm}
    \end{subfigure}
    \vspace{-8pt}
    \caption{The MPC policy with GP dynamics responding to 5 perturbations. The segway remains stable after each disturbance. \cref{fig:hw_gp_perturbation_inputs} shows the inputs spiking after each disturbance. The difference between inputs suggests the learned dynamics model captures asymmetries induced by the placement and sway of the unmodelled weights. \cref{fig:hw_mpc_perturbation_balanced} plots the angle of the segway with respect to the ground. \cref{fig:hw_mpc_perturbation_error_norm} shows the MPC's one step prediction error remains low through all disturbances.}
    \label{fig:hw_gp_perturbation}
\end{figure*}

\begin{figure*}[h!]
    \begin{subfigure}{0.32\linewidth}
    \hspace*{-10pt}
    \includegraphics[width=1.0\linewidth]{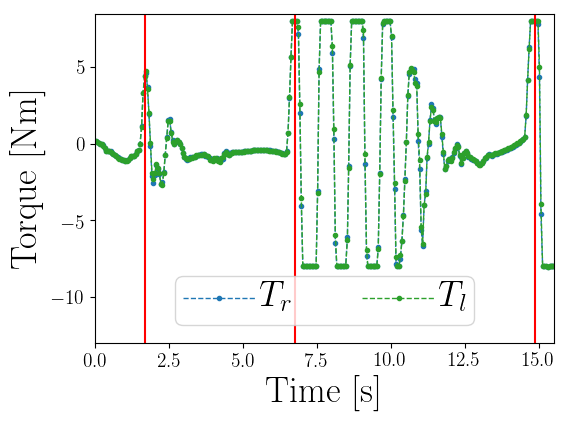}
    \vspace{-5pt}
    \caption{}
    \label{fig:hw_mpc_perturbation_inputs}
    \end{subfigure}
    \hspace*{-10pt}
    \begin{subfigure}{0.32\linewidth}
    \hspace*{-10pt}
    \includegraphics[width=1.0\linewidth]{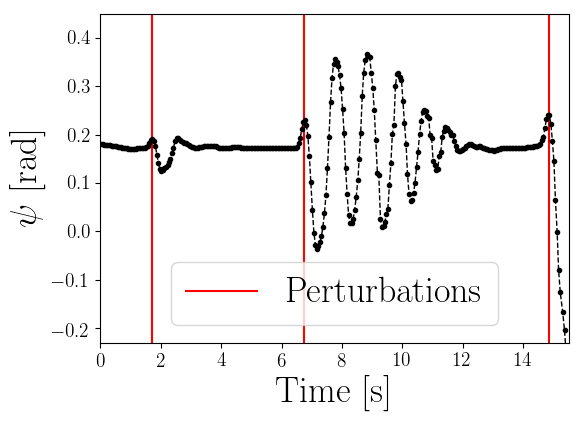}
    \vspace{-5pt}
    \caption{}
    \label{fig:hw_mpc_perturbation_balanced}
    \end{subfigure}
    \hspace*{-8pt}
    \begin{subfigure}{0.32\linewidth}
    \hspace*{-10pt}
    \includegraphics[width=1.0\linewidth]{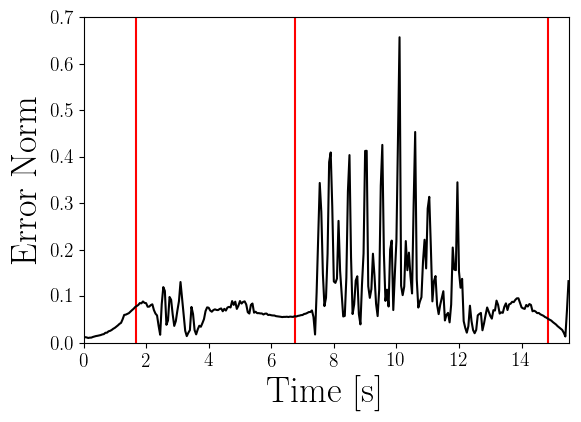}
    \vspace{-5pt}
    \caption{}
    \label{fig:hw_mpc_perturbation_error_norm}
    \end{subfigure}
    \vspace{-8pt}
    \caption{The MPC policy with nominal dynamics responding three perturbations. The segway remains stable after the first disturbance, oscillates before stabilizing for the second disturbance and falls down after the third disturbance. \cref{fig:hw_gp_perturbation_inputs} shows the input spike on each input after the disturbances. Notice that for all three perturbations, the both motors act in unison to stabilize the system. \cref{fig:hw_mpc_perturbation_balanced} shows the angle of the segway with respect to the ground. \cref{fig:hw_mpc_perturbation_error_norm} shows the MPC's one-step prediction error using the nominal dynamics. Notice that the oscillations cause the weight to swing which magnify the prediction error.}
    \label{fig:hw_mpc_perturbation}
\end{figure*}

In \cref{fig:hw_gp_perturbation} and \cref{fig:hw_mpc_perturbation}, we can see the result of applying force perpendicular to the axis of the wheels to the MPC policy with the nominal and GP dynamics, respectively. Both controllers have a spike in input following each disturbance, and in both cases the control action is saturated. Notice that because of the symmetry of the nominal model, the MPC policy applies the same force on each input as shown in \cref{fig:hw_mpc_perturbation_inputs}. Meanwhile the learned dynamics captures some of the asymmetry resulting from the weights which causes uneven outputs and a more robust system. In \cref{fig:hw_gp_perturbation_balanced} and \cref{fig:hw_mpc_perturbation_balanced}, we can see that even though the initial disturbances are of similar magnitude, the controller with nominal dynamics exhibits much larger oscillations and falls after the third perturbation. Although both models have sharp increases in one-step prediction error after a disturbance, the MPC model reaches much higher one-step prediction errors than with the GP, as shown in \cref{fig:hw_gp_perturbation_error_norm} and \cref{fig:hw_mpc_perturbation_error_norm}.



\section{Discussion and Future Work}
We presented a methodology for full dynamics learning that has been validated on an open-loop unstable robotic system. Using one minute of highly correlated data we are able estimate an accurate enough model for motion planning that is resilient to perturbations. Furthermore, the results presented in this paper are for a worst-case scenario where no useful prior is provided.  Finally, our approach is generic and can be applied to many Gaussian process modeling approaches as a drop-in subroutine.

There are many directions for future work.  A natural one is to study even higher-dimensional systems where one would likely need to combine learning with a prior nominal model.  Another direction is dealing with noise in the state estimation as well as delays, which are significant issues for unstable dynamical systems. Using techniques to correct noisy estimation data would significantly improve the performance of our method in real systems. This would be  true when dealing with outlying measurements that strongly violate the Gaussian assumption implicit in the GP.  One could also consider integration with perceptual systems \cite{dean2020robust}.


Another direction for future work is how to intelligently and autonomously collect training data.  A relevant line of work here is the area of safe exploration
\cite{liu_robust_2019,nakka2020chance,koller2018learning,sui2015safe,wachi2018safe,turchetta2019safe}.  It is also important to understand the fundamental limits of how much data we need to learn a reliable model, a concept known as
sample complexity in the machine learning literature \cite{wabersich2020bayesian}.

This work exploited the differentiability of Gaussian processes, but largely ignored the uncertainty quantification aspect.  In cases where there are more complex constraints to be satisfied, such as reachability \cite{akametalu2014reachability} or chance constraints
\cite{nakka2020chance}, it would be interesting to develop a more holistic framework that reasons about uncertainty quantication in differentiated GP models.

A final direction for future work is scalability.  For more complex systems, it would be beneficial to collected and store more data points for estimating the GP model.  However, it is known that the computational complexity of GP inference can scale poorly with the amount of training data.  Leveraging various methods for scaling up GP training and inference could be beneficial \cite{wilson2015kernel,snelson2007local}.

\section*{Acknowledgments}
We would like to thank Andrew Singletary and Ellen Novoseller for their help in setting up the robotics and the mathematical derivations, respectively. We would also like to thank the follwing software packages: PyTorch \cite{NEURIPS2019_9015}, CVXPY \cite{agrawal2018rewriting} and Gurobi \cite{gurobi}.


\bibliographystyle{IEEEtran}
\renewcommand{\baselinestretch}{0.91}
\bibliography{references.bib}

\begin{thebibliography}{10}
\providecommand{\url}[1]{#1}
\csname url@samestyle\endcsname
\providecommand{\newblock}{\relax}
\providecommand{\bibinfo}[2]{#2}
\providecommand{\BIBentrySTDinterwordspacing}{\spaceskip=0pt\relax}
\providecommand{\BIBentryALTinterwordstretchfactor}{4}
\providecommand{\BIBentryALTinterwordspacing}{\spaceskip=\fontdimen2\font plus
\BIBentryALTinterwordstretchfactor\fontdimen3\font minus
  \fontdimen4\font\relax}
\providecommand{\BIBforeignlanguage}[2]{{%
\expandafter\ifx\csname l@#1\endcsname\relax
\typeout{** WARNING: IEEEtran.bst: No hyphenation pattern has been}%
\typeout{** loaded for the language `#1'. Using the pattern for}%
\typeout{** the default language instead.}%
\else
\language=\csname l@#1\endcsname
\fi
#2}}
\providecommand{\BIBdecl}{\relax}
\BIBdecl

\bibitem{lai1982least}
T.~L. Lai, C.~Z. Wei \emph{et~al.}, ``Least squares estimates in stochastic
  regression models with applications to identification and control of dynamic
  systems,'' \emph{Annals of Statistics}, vol.~10, no.~1, pp. 154--166, 1982.

\bibitem{khosla_parameter_1985}
P.~K. Khosla and T.~Kanade, ``Parameter identification of robot dynamics,'' in
  \emph{1985 24th {IEEE} Conference on Decision and Control}, pp. 1754--1760.

\bibitem{chen1990non}
S.~Chen, S.~Billings, and P.~Grant, ``Non-linear system identification using
  neural networks,'' \emph{International journal of control}, vol.~51, no.~6,
  pp. 1191--1214, 1990.

\bibitem{caydas_support_2012}
\BIBentryALTinterwordspacing
U.~Çaydaş and S.~Ekici, ``Support vector machines models for surface
  roughness prediction in {CNC} turning of {AISI} 304 austenitic stainless
  steel,'' vol.~23, no.~3, pp. 639--650. [Online]. Available:
  \url{https://doi.org/10.1007/s10845-010-0415-2}
\BIBentrySTDinterwordspacing

\bibitem{willis_systems_nodate}
M.~Willis, H.~Hiden, M.~Hinchliffe, and B.~{McKAY}, ``Systems modelling using
  genetic programming,'' p.~6.

\bibitem{borrelli2017predictive}
F.~Borrelli, A.~Bemporad, and M.~Morari, \emph{Predictive control for linear
  and hybrid systems}.\hskip 1em plus 0.5em minus 0.4em\relax Cambridge
  University Press, 2017.

\bibitem{aswani2013provably}
A.~Aswani, H.~Gonzalez, S.~S. Sastry, and C.~Tomlin, ``Provably safe and robust
  learning-based model predictive control,'' \emph{Automatica}, vol.~49, no.~5,
  pp. 1216--1226, 2013.

\bibitem{papadimitriou_control_2020}
\BIBentryALTinterwordspacing
D.~Papadimitriou, U.~Rosolia, and F.~Borrelli, ``Control of unknown nonlinear
  systems with linear time-varying {MPC}.'' [Online]. Available:
  \url{http://arxiv.org/abs/2004.03041}
\BIBentrySTDinterwordspacing

\bibitem{koller2018learning}
T.~Koller, F.~Berkenkamp, M.~Turchetta, and A.~Krause, ``Learning-based model
  predictive control for safe exploration,'' in \emph{2018 IEEE Conference on
  Decision and Control (CDC)}.\hskip 1em plus 0.5em minus 0.4em\relax IEEE,
  2018, pp. 6059--6066.

\bibitem{rosolia2019learning}
U.~Rosolia and F.~Borrelli, ``Learning how to autonomously race a car: a
  predictive control approach,'' \emph{IEEE Transactions on Control Systems
  Technology}, vol.~28, no.~6, pp. 2713--2719, 2019.

\bibitem{hewing2019cautious}
L.~Hewing, J.~Kabzan, and M.~N. Zeilinger, ``Cautious model predictive control
  using gaussian process regression,'' \emph{IEEE Transactions on Control
  Systems Technology}, vol.~28, no.~6, pp. 2736--2743, 2019.

\bibitem{kabzan2019learning}
J.~Kabzan, L.~Hewing, A.~Liniger, and M.~N. Zeilinger, ``Learning-based model
  predictive control for autonomous racing,'' \emph{IEEE Robotics and
  Automation Letters}, vol.~4, no.~4, pp. 3363--3370, 2019.

\bibitem{klenske2015gaussian}
E.~D. Klenske, M.~N. Zeilinger, B.~Sch{\"o}lkopf, and P.~Hennig, ``Gaussian
  process-based predictive control for periodic error correction,'' \emph{IEEE
  Transactions on Control Systems Technology}, vol.~24, no.~1, pp. 110--121,
  2015.

\bibitem{shi2019neural}
G.~Shi, X.~Shi, M.~O’Connell, R.~Yu, K.~Azizzadenesheli, A.~Anandkumar,
  Y.~Yue, and S.-J. Chung, ``Neural lander: Stable drone landing control using
  learned dynamics,'' in \emph{2019 International Conference on Robotics and
  Automation (ICRA)}.\hskip 1em plus 0.5em minus 0.4em\relax IEEE, 2019, pp.
  9784--9790.

\bibitem{shi2020neural}
G.~Shi, W.~H{\"o}nig, Y.~Yue, and S.-J. Chung, ``Neural-swarm: Decentralized
  close-proximity multirotor control using learned interactions,'' in
  \emph{2020 IEEE International Conference on Robotics and Automation
  (ICRA)}.\hskip 1em plus 0.5em minus 0.4em\relax IEEE, 2020, pp. 3241--3247.

\bibitem{taylor2020learning}
A.~Taylor, A.~Singletary, Y.~Yue, and A.~Ames, ``Learning for safety-critical
  control with control barrier functions,'' in \emph{Learning for Dynamics and
  Control}.\hskip 1em plus 0.5em minus 0.4em\relax PMLR, 2020, pp. 708--717.

\bibitem{taylor2019episodic}
A.~J. Taylor, V.~D. Dorobantu, H.~M. Le, Y.~Yue, and A.~D. Ames, ``Episodic
  learning with control lyapunov functions for uncertain robotic systems,'' in
  \emph{2019 IEEE/RSJ International Conference on Intelligent Robots and
  Systems (IROS)}.\hskip 1em plus 0.5em minus 0.4em\relax IEEE, 2019, pp.
  6878--6884.

\bibitem{chang2017learning}
A.~H. Chang, C.~M. Hubicki, J.~J. Aguilar, D.~I. Goldman, A.~D. Ames, and P.~A.
  Vela, ``Learning to jump in granular media: Unifying optimal control
  synthesis with gaussian process-based regression,'' in \emph{2017 IEEE
  International Conference on Robotics and Automation (ICRA)}.\hskip 1em plus
  0.5em minus 0.4em\relax IEEE, 2017, pp. 2154--2160.

\bibitem{bujarbaruah2018adaptive}
M.~Bujarbaruah, X.~Zhang, U.~Rosolia, and F.~Borrelli, ``Adaptive mpc for
  iterative tasks,'' in \emph{2018 IEEE Conference on Decision and Control
  (CDC)}.\hskip 1em plus 0.5em minus 0.4em\relax IEEE, 2018, pp. 6322--6327.

\bibitem{deisenroth_gaussian_2015}
\BIBentryALTinterwordspacing
M.~P. Deisenroth, D.~Fox, and C.~E. Rasmussen, ``Gaussian processes for
  data-efficient learning in robotics and control,'' vol.~37, no.~2, pp.
  408--423. [Online]. Available:
  \url{http://ieeexplore.ieee.org/document/6654139/}
\BIBentrySTDinterwordspacing

\bibitem{hewing_cautious_2020}
\BIBentryALTinterwordspacing
L.~Hewing, J.~Kabzan, and M.~N. Zeilinger, ``Cautious model predictive control
  using gaussian process regression,'' vol.~28, no.~6, pp. 2736--2743.
  [Online]. Available: \url{https://ieeexplore.ieee.org/document/8909368/}
\BIBentrySTDinterwordspacing

\bibitem{wang2018safe}
L.~Wang, E.~A. Theodorou, and M.~Egerstedt, ``Safe learning of quadrotor
  dynamics using barrier certificates,'' in \emph{2018 IEEE International
  Conference on Robotics and Automation (ICRA)}.\hskip 1em plus 0.5em minus
  0.4em\relax IEEE, 2018, pp. 2460--2465.

\bibitem{kocijan2004gaussian}
J.~Kocijan, R.~Murray-Smith, C.~E. Rasmussen, and A.~Girard, ``Gaussian process
  model based predictive control,'' in \emph{Proceedings of the 2004 American
  control conference}, vol.~3.\hskip 1em plus 0.5em minus 0.4em\relax IEEE,
  2004, pp. 2214--2219.

\bibitem{mchutchon2013differentiating}
\BIBentryALTinterwordspacing
A.~McHutchon, ``Differentiating gaussian processes,'' 2013. [Online].
  Available: \url{http://mlg.eng.cam.ac.uk/mchutchon/DifferentiatingGPs.pdf}
\BIBentrySTDinterwordspacing

\bibitem{solak_derivative_nodate}
E.~Solak, R.~Murray-smith, W.~E. Leithead, D.~J. Leith, and C.~E. Rasmussen,
  ``Derivative observations in gaussian process models of dynamic systems,''
  p.~8.

\bibitem{lee_gp-ilqg_2017}
\BIBentryALTinterwordspacing
G.~Lee, S.~S. Srinivasa, and M.~T. Mason, ``{GP}-{ILQG}: Data-driven robust
  optimal control for uncertain nonlinear dynamical systems.'' [Online].
  Available: \url{http://arxiv.org/abs/1705.05344}
\BIBentrySTDinterwordspacing

\bibitem{levine2021framework}
M.~E. Levine and A.~M. Stuart, ``A framework for machine learning of model
  error in dynamical systems,'' 2021.

\bibitem{pan_probabilistic_nodate}
Y.~Pan and E.~Theodorou, ``Probabilistic differential dynamic programming,''
  p.~9.

\bibitem{deisenroth_pilco_nodate}
M.~P. Deisenroth and C.~E. Rasmussen, ``{PILCO}: A model-based and
  data-efficient approach to policy search,'' p.~8.

\bibitem{lutter_value_2021}
\BIBentryALTinterwordspacing
M.~Lutter, S.~Mannor, J.~Peters, D.~Fox, and A.~Garg, ``Value iteration in
  continuous actions, states and time.'' [Online]. Available:
  \url{http://arxiv.org/abs/2105.04682}
\BIBentrySTDinterwordspacing

\bibitem{hewing_learning-based_2020}
\BIBentryALTinterwordspacing
L.~Hewing, K.~P. Wabersich, M.~Menner, and M.~N. Zeilinger, ``Learning-based
  model predictive control: Toward safe learning in control,'' vol.~3, no.~1,
  pp. 269--296, \_eprint:
  https://doi.org/10.1146/annurev-control-090419-075625. [Online]. Available:
  \url{https://doi.org/10.1146/annurev-control-090419-075625}
\BIBentrySTDinterwordspacing

\bibitem{quinonero2005unifying}
J.~Quinonero-Candela and C.~E. Rasmussen, ``A unifying view of sparse
  approximate gaussian process regression,'' \emph{The Journal of Machine
  Learning Research}, vol.~6, pp. 1939--1959, 2005.

\bibitem{dean2020robust}
S.~Dean, N.~Matni, B.~Recht, and V.~Ye, ``Robust guarantees for
  perception-based control,'' in \emph{Learning for Dynamics and
  Control}.\hskip 1em plus 0.5em minus 0.4em\relax PMLR, 2020, pp. 350--360.

\bibitem{liu_robust_2019}
\BIBentryALTinterwordspacing
A.~Liu, G.~Shi, S.-J. Chung, A.~Anandkumar, and Y.~Yue, ``Robust regression for
  safe exploration in control.'' [Online]. Available:
  \url{http://arxiv.org/abs/1906.05819}
\BIBentrySTDinterwordspacing

\bibitem{nakka2020chance}
Y.~K. Nakka, A.~Liu, G.~Shi, A.~Anandkumar, Y.~Yue, and S.-J. Chung,
  ``Chance-constrained trajectory optimization for safe exploration and
  learning of nonlinear systems,'' \emph{IEEE Robotics and Automation Letters},
  vol.~6, no.~2, pp. 389--396, 2020.

\bibitem{sui2015safe}
Y.~Sui, A.~Gotovos, J.~Burdick, and A.~Krause, ``Safe exploration for
  optimization with gaussian processes,'' in \emph{International Conference on
  Machine Learning}.\hskip 1em plus 0.5em minus 0.4em\relax PMLR, 2015, pp.
  997--1005.

\bibitem{wachi2018safe}
A.~Wachi, Y.~Sui, Y.~Yue, and M.~Ono, ``Safe exploration and optimization of
  constrained mdps using gaussian processes,'' in \emph{Proceedings of the AAAI
  Conference on Artificial Intelligence}, vol.~32, no.~1, 2018.

\bibitem{turchetta2019safe}
M.~Turchetta, F.~Berkenkamp, and A.~Krause, ``Safe exploration for interactive
  machine learning,'' \emph{arXiv preprint arXiv:1910.13726}, 2019.

\bibitem{wabersich2020bayesian}
K.~P. Wabersich and M.~Zeilinger, ``Bayesian model predictive control:
  Efficient model exploration and regret bounds using posterior sampling,'' in
  \emph{Learning for Dynamics and Control}.\hskip 1em plus 0.5em minus
  0.4em\relax PMLR, 2020, pp. 455--464.

\bibitem{akametalu2014reachability}
A.~K. Akametalu, J.~F. Fisac, J.~H. Gillula, S.~Kaynama, M.~N. Zeilinger, and
  C.~J. Tomlin, ``Reachability-based safe learning with gaussian processes,''
  in \emph{53rd IEEE Conference on Decision and Control}.\hskip 1em plus 0.5em
  minus 0.4em\relax IEEE, 2014, pp. 1424--1431.

\bibitem{wilson2015kernel}
A.~Wilson and H.~Nickisch, ``Kernel interpolation for scalable structured
  gaussian processes (kiss-gp),'' in \emph{International Conference on Machine
  Learning}.\hskip 1em plus 0.5em minus 0.4em\relax PMLR, 2015, pp. 1775--1784.

\bibitem{snelson2007local}
E.~Snelson and Z.~Ghahramani, ``Local and global sparse gaussian process
  approximations,'' in \emph{Artificial Intelligence and Statistics}.\hskip 1em
  plus 0.5em minus 0.4em\relax PMLR, 2007, pp. 524--531.

\bibitem{NEURIPS2019_9015}
A.~Paszke, S.~Gross, F.~Massa, A.~Lerer, J.~Bradbury, G.~Chanan, T.~Killeen,
  Z.~Lin, N.~Gimelshein, L.~Antiga, A.~Desmaison, A.~Kopf, E.~Yang, Z.~DeVito,
  M.~Raison, A.~Tejani, S.~Chilamkurthy, B.~Steiner, L.~Fang, J.~Bai, and
  S.~Chintala, ``Pytorch: An imperative style, high-performance deep learning
  library,'' in \emph{Advances in Neural Information Processing Systems 32},
  H.~Wallach, H.~Larochelle, A.~Beygelzimer, F.~d\textquotesingle
  Alch\'{e}-Buc, E.~Fox, and R.~Garnett, Eds.\hskip 1em plus 0.5em minus
  0.4em\relax Curran Associates, Inc., 2019, pp. 8024--8035.

\bibitem{agrawal2018rewriting}
A.~Agrawal, R.~Verschueren, S.~Diamond, and S.~Boyd, ``A rewriting system for
  convex optimization problems,'' \emph{Journal of Control and Decision},
  vol.~5, no.~1, pp. 42--60, 2018.

\bibitem{gurobi}
\BIBentryALTinterwordspacing
{Gurobi Optimization, LLC}, ``{Gurobi Optimizer Reference Manual},'' 2021.
  [Online]. Available: \url{https://www.gurobi.com}
\BIBentrySTDinterwordspacing

\end{thebibliography}

\end{document}